# Automatic Recognition and Digital Documentation of Cultural Heritage Hemispherical Domes using Images

Reza Maalek*

Chair of Department of Digital Engineering and Construction (DEC), Karlsruhe Institute of Technology (KIT), reza.maalek@kit.edu (corresponding author)

Shahrokh Maalek

Chief Technology Officer (CTO), Digital Information in Construction Engineering (DICE) Technologies, shahrokh.maalek@dicetechnologies.ca

## ABSTRACT

Recent advancements in optical metrology have enabled continuous documentation of dense 3-dimensional (3D) point clouds of construction projects, including cultural heritage preservation projects. These point clouds must then be further processed to generate semantic digital models, which is integral to the lifecycle management of heritage sites. For large scale and continuous digital documentation, processing of dense 3D point clouds is computationally cumbersome, and consequentially requires additional hardware for data management and analysis, increasing the time, cost, and complexity of the project. Fast and reliable solutions for generating the geometric digital models is, hence, eminently desirable. This manuscript presents an original approach to generate reliable semantic digital models of heritage hemispherical domes using only two images. New closed formulations were derived to establish the relationships between a sphere and its projected ellipse onto an image. These formulations were then utilized to create new methods for: (i) selecting the best pair of images from an image network; (ii) detecting ellipses corresponding to projection of spheres in images; (iii) matching of the detected ellipses between images; and (iv) generating the sphere's geometric digital models. The effectiveness of the proposed method was evaluated under both laboratory and real-world datasets. Laboratory experiments revealed that the proposed process using the best pair of images provided results as accurate as that achieved using eight randomly selected images, while improving computation time by a factor of 50. The results of the two real-world datasets showed that the digital model of a hemispherical dome was generated with 6.2mm accuracy, while improving the total computation time of current best practice by a factor of 7. Real-world experimentation also showed that the proposed method can provide metric scale definition for photogrammetric point clouds with 3mm accuracy using spherical targets. The results suggest that the proposed method was successful in automatically generating fast, and accurate geometric digital models of hemispherical domes.

* Corresponding author.

# 1 Introduction and background

## 1.1 Significance of digital documentation for heritage hemispherical domes

Hemispherical domes are common geometric elements found in the historical developments of many civilizations, including the Persians, Greeks, Romans, and Assyrians [12,34]. Figure 1a shows an artistic impression of typical hemispherical domes in urban developments pertaining to the surroundings of the central Iranian desert (Kavir) dating back to the medieval era. Such old cities still exist, accommodate a considerable population, and in some cases -due to their cultural and historical values as well as their urban planning and architectural aspects- are recognized as important cultural heritage sites by the UNESCO [39]. We have witnessed the destruction of such cultural heritage sites many times in history as a result of either natural disasters, such as the earthquake in Bam [3], or man-made disasters, such as wars. Let's assume that as a consequence of such disasters, the front part of the city is destroyed. If an up-to-date digital twin model [4,15] of these domes exists (Figure 1b), the destructed part can be revived and redeveloped to its original form – e.g., using a combination of computational topological optimization methods and lightweight spatial structure [26] construction techniques – as schematically depicted in Figure 1c. In fact, the lightweight reconstruction of the outer shell of heritage domes has also shown to be instrumental in minimizing long-term and excessive building settlements [11]. Hence, the first step to create a workable and up-to-date digital twin for sustainable and lightweight redevelopment is to propose a solution for frequent 3D as-built digital documentation of these domes, specifically the outer shell, which will be the primary focus of this study. The as-built 3D model together with the material properties (transfer and mechanical properties) and other relevant information can then be used to generate digital twins of these domes.

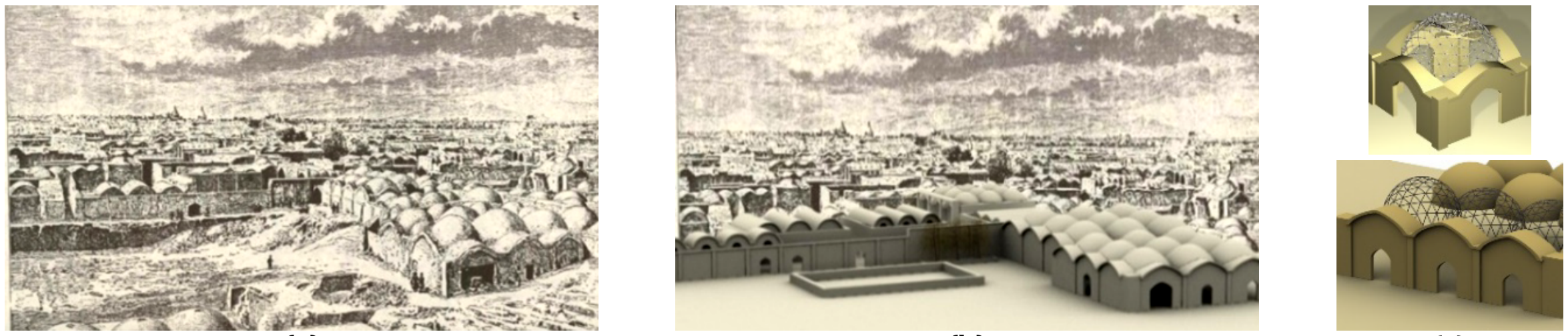

(a) (b) (c)

**Figure 1: Typical hemispherical domes (Gonbad) in the vicinity of the central Iranian dessert (Kavir): (a) artistic impression of Koum by Madam Jane Dieulafoy [7]; (b) superimposed digital as-built model; and (c) spatial structure for the optimized redevelopment strategy of the digitally documented heritage dome.**

## 1.2 As-built digital documentation of hemispherical domes using point clouds

Recent advancements in photogrammetry [1,33], and laser scanning has enabled the collection of accurate 3-dimensional (3D) point clouds of cultural heritage sites [28,30,38]. The coordinates of the points -and where appropriate color or intensity information- must then be utilized to detect and generate the semantic digital models of these hemispherical domes [15,19,24,27]. Figure 2 illustrates a typical process for the semantic digital model generation of a hemispherical dome using a network of overlapping images [19]. To acquire the point cloud, a process, called structure-from-motion (SfM) [35], is commonly adopted. SfM starts by automatically detecting and matching scale-invariant features [14] between different images so as to determine the exterior orientation parameters (EOPs) of each image (position and orientations shown with red prisms in Figure 2a), along with the camera's interior orientation parameters (IOPs) [9]. The EOPs and IOPs from the SfM are then used to generate dense 3D point clouds (Figure 2b). The dense point clouds can of course also be acquired through other means, such as laser scanning. The points representing the hemispherical domes must then be segmented and modeled to generate the as-built digital model (red in Figure 2c) [19,24,27]. The latter can be performed either manually, or automatically. Manual processing of point clouds is time consuming [15] and subject to human-error [19,24,36]. Therefore, recent research has focused on the development of solutions to automatically process point clouds and generate semantic digital models. The latter is sometimes referred to as scan-to-building information model (BIM) in the construction and building informatics literature [5].

While much research has been successfully carried out to provide solutions for automatic generation of semantic as-built digital models from point clouds [15,18,24,31,32,36], the methods can become computationally expensive, the faster heuristic processing methods are typically biased to specific types of scenes or noise levels, and supervised learning methods require a large library of pre-classified point clouds [17,18,24]. As such, this manuscript formulates the scientific bases to propose a robust solution for as-built digital modeling of hemispherical domes at a fraction of the computation time of reliable point cloud processing methods using only one pair of images. The method is extendable to the detection and modeling of any spherical object, and consequentially can also be utilized for other important applications, such as calibration and metric scale definition of photogrammetric point clouds using spherical targets.

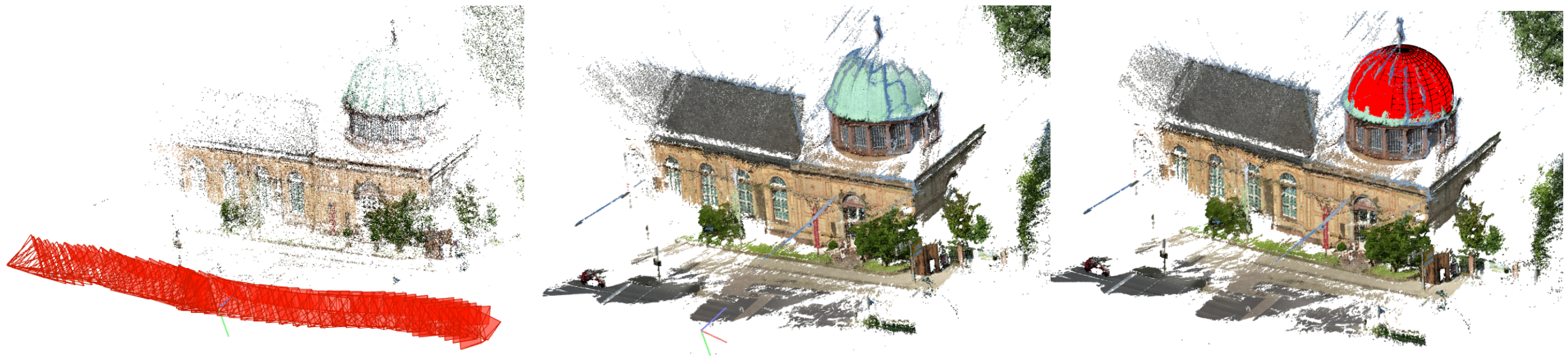

(a) (b) (c)

**Figure 2: As-built modeling of hemispherical domes [19,29]: (a) coarse 3D reconstruction; (b) dense 3D reconstructed point cloud; and (c) 3D as-built model of the hemispherical dome in red.**

## 1.3 Manuscript objective and scope

The main research aim is to develop a solution for large-scale semantic 3D digital documentation and model generation of hemispherical domes. This study provides a simple, reliable, and computationally efficient original solution for detecting and semantic 3D modeling of these domes using only a pair of images. To this end, new methods were developed for: (i) closed formulation of the projection of spheres onto images; (ii) detection of spherical ellipses in images; (iii) matching of the detected ellipses between images; (iv) modeling of the parent sphere using the set of matched ellipses between the images; and (v) detection of the best pair of images from a network of overlapping images. The study does not consider the detection and digital modeling of non-hemispherical or geometrically damaged hemispherical domes.

# 2 Methodology

In this section, the proposed method to achieve the objectives outlined in Section 1.3 are presented. The section is divided into two sub-categories, namely, “closed formulation of sphere reconstruction in images”, and “automatic as-build modeling of spherical structures”. In the first sub-category, a set of closed formulations are derived to represent the relationships between spheres and the corresponding projected ellipse onto one image. Using these formulations, a set of mathematical corollaries and propositions are made to establish the necessary and sufficient conditions to reconstruct object space spheres using ellipses in images. These findings are then combined into an algorithm, which inputs matching ellipses from two or more images, to generate the as-built model of the parent spheres. The second sub-category focuses on the full automation of the as-built modeling process developed in the former sub-category using only a pair of images. As such, the newly developed process for the automatic detection and matching of spherical ellipses between images, along with the method of selection of the best pair of images is explained.

## 2.1 Closed formulation of sphere reconstruction in images

The projection of a sphere onto an image is an ellipse. This is since the sphere and the camera’s center create a cone, whereby the cone’s vertex and base are the camera center and locus of points tangential to the sphere, respectively. The intersection of this cone with the intersecting image plane forms an ellipse. To provide a visual perspective, Figure 3a depicts a typical sphere, the cone that is tangential to the sphere, the plane intersecting the cone and the projected ellipse. In this section, the goal is to establish the closed form relationships between the original sphere parameters (center and radius) and the projected ellipse parameters, given the camera projective matrix (IOPs and EOPs). The ellipse parameters $\rho = (C_e, a_e, b_e, \theta)$ are the center, semi-major length, semi-minor length, and the angle between the major axis and the horizontal axis [21].

### 2.1.1 Preliminaries: camera projective matrix

The camera projective matrix, $P_J$, establishes the relationship between a point in space and its projection onto an image [9]. For the case shown in Figure 3b, the relationship between a point $C$ and its projection $C_C$ is $C_C = P_J C$. In the camera coordinate frame, i.e., $X_{cam}$-$Y_{cam}$-$Z_{cam}$, the relationship can be formulated using the camera’s focal length, $f$, and principal point, $P = \left(p_x, p_y\right)^T$, as depicted in Figure 3b-right:

$$C_C = \begin{bmatrix} f & 0 & p_x & 0 \\ 0 & f & p_y & 0 \\ 0 & 0 & 1 & 0 \end{bmatrix} C = \begin{bmatrix} f & 0 & p_x \\ 0 & f & p_y \\ 0 & 0 & 1 \end{bmatrix} [I_{3\times3}|0_{3\times1}]C = K[I|0]C \rightarrow P_J = K[I|0] \qquad (1)$$

where $K$ is referred to as the camera calibration matrix, representing the IOPs. If the world coordinate system is represented in the coordinate system of $O'$, where $O$ and $O'$ are separated by a rigid body transformation with rotation, $Rot$, and translation, $t$, the camera projective matrix changes to the general form of: $P_J = K[Rot_{3\times3}|t_{3\times1}]$. The camera coordinate frame is a coordinate system where the camera center, $O$, is at the origin and the image plane’s normal (principal axis) is parallel to the $Z$ axis. To transform the world coordinate system to the camera coordinate frame, one must, hence, use the transformation matrix $Rt = \begin{bmatrix} Rot & t \\ 0 & 1 \end{bmatrix}$, representative of the EOPs. For ease of formulation and without loss of generality, the equations derived in the following are performed in the camera coordinate frame.

### 2.1.2 Projected ellipse parameters from original sphere

Using tangent balls [8], Dandelin showed that the intersection of a plane and a cone produces an ellipse. The centers of the two tangential Dendelin spheres (see Figure 3c), and the major axis of the generated ellipse, lie on a plane ($\Delta O M_{C1} M_{C2}$) whose normal vector is parallel to the minor axis of the ellipse. Figure 3c shows the arrangement of Figure 3a together with the Dandelin’s spheres, projected onto the plane $\Delta O M_{C1} M_{C2}$. It can be observed that the line segment $\overline{OC}$ passes through $C_C$ and the centers of the Dandelin spheres, $D_1$ and $D_2$. The projection of the line segment $\overline{OC}$ onto the image plane represents the major axis direction of the projected ellipse. This is since the projection of $D_1$ and $D_2$ onto the image plane are the two foci of the projected ellipse, $F_1$ and $F_2$, respectively. By extension, the major axis also passes through the projection of the camera center, $O$, onto the image plane, $P$. Therefore, the camera's perspective center, $P$, lies on the major axis of the projected ellipse. Using this information together with the formulations provided in Figure 3b and eq. 1, the direction of the major axis of the ellipse, $\overrightarrow{M_J}$, shown in Figure 3d, and the angle between the major axis and the horizontal axis, $\theta$, are determined as follows:

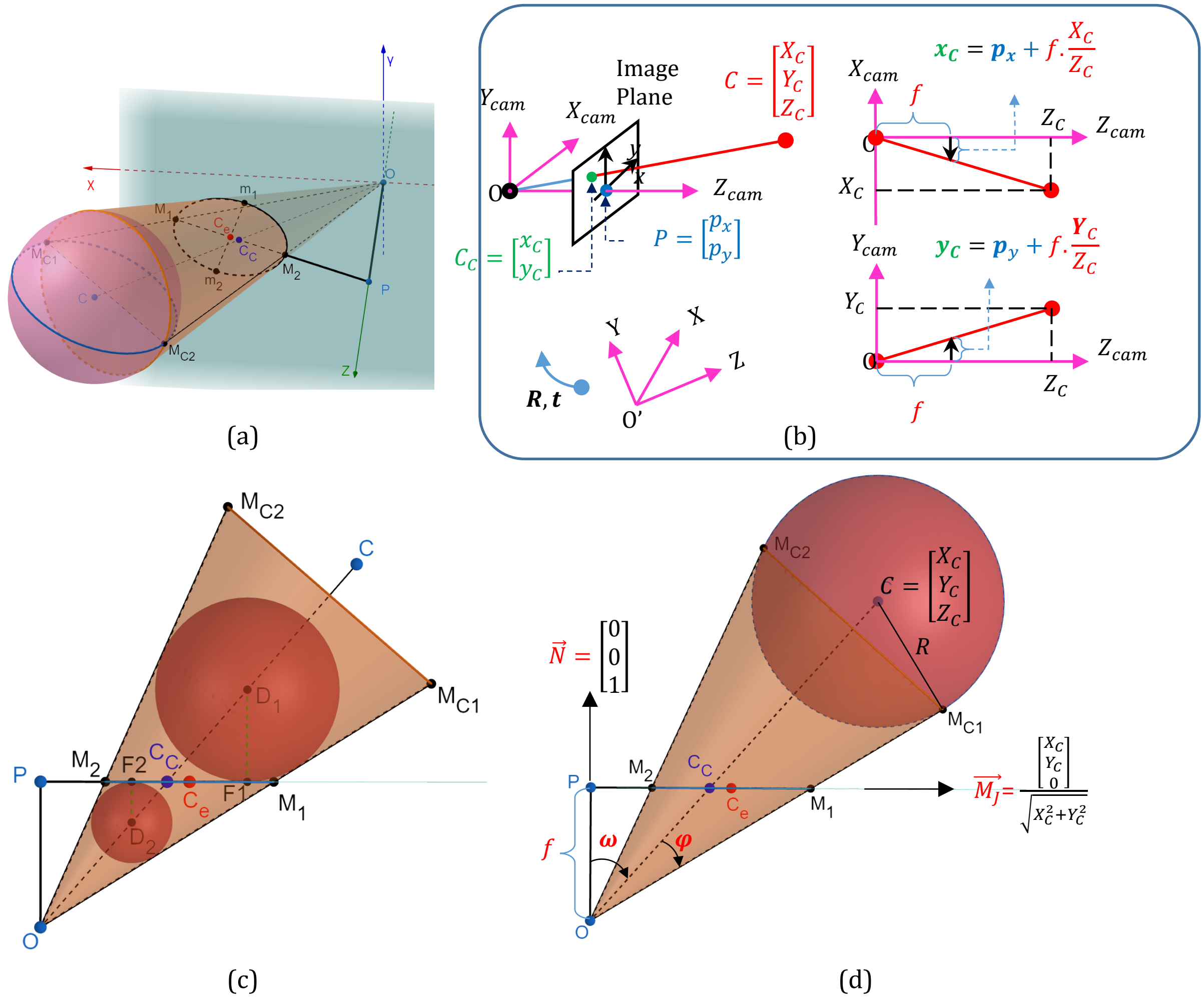

| | | | |
|---|---|---|---|
| $C$: sphere center | $C_C$: projection of $C$ onto image plane | $C_e$: projected ellipse's center | $O$: camera center |
| $M_{Ci}$: tangents from $O$ to the sphere ($i = 1,2$) | | $M_i$: ellipse major points | $P$: camera's principal point |
| $R$: sphere radius | $f$: camera focal length | $M_J$: major axis of ellipse | $N$: image plane normal |

**Figure 3: Schematics of the projection of spheres onto images: a) 3D orthographic visual; b) representation of point projection in the pinhole camera model; c) Dandelin's spheres; and d) sphere's projection.**

$$\begin{cases} C_C = \begin{bmatrix} p_x + f.\frac{X_C}{Z_C} \\ p_y + f.\frac{Y_C}{Z_C} \\ f \end{bmatrix}, \quad P = \begin{bmatrix} p_x \\ p_y \\ f \end{bmatrix} \\ \overrightarrow{M_J} = \frac{\overrightarrow{PC_C}}{|\overrightarrow{PC_C}|} = \frac{C_C - P}{|C_C - P|} = \frac{\frac{f}{Z_C}\begin{bmatrix} X_C \\ Y_C \\ 0 \end{bmatrix}}{\frac{f}{Z_C}\sqrt{X_C^2 + Y_C^2}} = \frac{\begin{bmatrix} X_C \\ Y_C \\ 0 \end{bmatrix}}{\sqrt{X_C^2 + Y_C^2}} \rightarrow \theta = \arctan\frac{Y_C}{X_C} \end{cases} \tag{2}$$

where $C = (X_C, Y_C, Z_C)$ is the center of the sphere in the camera's coordinate frame (Figure 3d). Using the major points, $M_1$ and $M_2$, the semi-major length of the ellipse, $a_e$, and the ellipse center, $C_e$, are calculated as follows:

$$\begin{cases} a_e = \frac{|M_1 - M_2|}{2} \\ C_e = \frac{M_1 + M_2}{2} \end{cases} \tag{3}$$

Using the triangles $\Delta OPM_2$, and $\Delta OPM_1$, line segments $\overline{PM_2}$ and $\overline{PM_1}$ are obtained as follows:

$$\begin{cases} \overline{PM_2} = f.\tan(\omega - \varphi) = f.\frac{\tan\omega - \tan\varphi}{1 + \tan\omega\tan\varphi} \\ \overline{PM_1} = f.\tan(\omega + \varphi) = f.\frac{\tan\omega + \tan\varphi}{1 - \tan\omega\tan\varphi} \end{cases} \tag{4}$$

Using the triangles $\Delta OM_{C1}C$, and $\Delta OPC_C$, together with eq. 2, the angles $\varphi$ and $\omega$, respectively, are calculated:

$$\begin{cases} \tan\varphi = \frac{R}{\sqrt{X_C^2 + Y_C^2 + Z_C^2 - R^2}} \\ \tan\omega = \frac{\sqrt{X_C^2 + Y_C^2}}{Z_C} \end{cases} \tag{5}$$

Substituting eq. 5 into 4 and then into 3 will provide the following simplified formulas:

$$\begin{cases} a_e = \frac{fR\sqrt{X_C^2 + Y_C^2 + Z_C^2 - R^2}}{Z_C^2 - R^2} \\ C_e = \begin{bmatrix} p_x + f.\frac{Z_C X_C}{Z_C^2 - R^2} \\ p_y + f.\frac{Z_C Y_C}{Z_C^2 - R^2} \end{bmatrix} \end{cases} \tag{6}$$

The last consideration is the semi-minor length, which is derived directly from the formulations presented in [8] using the Dendelin spheres (Figure 3c) as follows:

$$b_e = \frac{f\sin\varphi}{\sqrt{\cos^2\omega - \sin^2\varphi}} = \frac{fR}{\sqrt{Z_C^2 - R^2}} \tag{7}$$

From eq. 1, eq. 6 and eq. 7, one proposition and three corollaries are derived, which are explained in the following.

**Proposition 1.** The coordinates of the center of one sphere in the camera coordinate system can be derived using a single image of the sphere from a calibrated camera up to an arbitrary scale factor.

**Proof.** Since the camera is assumed calibrated, the camera calibration parameters, $f$, $p_x$ and $p_y$ are known. In practice, camera calibration can be performed using laboratory pre-calibration practices [20]. Since the projection of a sphere onto an image is an ellipse, a reliable method for ellipse detection from images [22] can be deployed to determine the ellipse's geometric parameters $\rho = (C_e, a_e, b_e, \theta)$. From eq. 7, the following relationships are derived:

$$Z_C = \frac{R\sqrt{f^2 + b_e^2}}{b_e} \tag{8}$$

Substituting $Z_C$ from eq. 8 into $C_e = (x_{Ce}, y_{Ce})$ in eq. 6 provides:

$$\begin{cases} X_C = \frac{fR}{b_e\sqrt{f^2+b_e^2}}(x_{Ce} - p_x) \\ Y_C = \frac{fR}{b_e\sqrt{f^2+b_e^2}}(y_{Ce} - p_y) \end{cases} \quad (9)$$

Therefore, $X_C$, $Y_C$ and $Z_C$ can be determined as a function of $f$, $P$, $C_e$ and $b_e$ and linearly dependent on a scale factor equal to $R$.

**Corollary 1.** $X_C$, $Y_C$ and $Z_C$ are derived in real-world scale if the real-world radius of the sphere is known.

**Proof.** The proof follows from the correlation of $X_C$, $Y_C$ and $Z_C$ with, $R$, in eq. 8 and eq. 9.

**Corollary 2.** The radius of a sphere is determined using a single image of the sphere from a calibrated camera if and only if the distance of the center of the sphere, projected onto the image's principal axis in the camera's coordinate frame, $Z_C$, is known.

**Proof.** Re-arranging eq. 8 gives:

$$R = \frac{Z_C . b_e}{\sqrt{b_e^2+f^2}} \quad (10)$$

Similarly, eq. 9 may also be used to estimate the radius of the sphere if either of $X_C$, or $Y_C$ in the camera's coordinate frame were known. The radii calculated using the latter formulations, however, become indeterminate when $x_{Ce} \to p_x$ or $y_{Ce} \to p_y$. On the other hand, eq. 10 suggests that knowledge of $Z_C$ provides the necessary and sufficient conditions to calculate the radius since the function is deterministic in $\mathbb{R}^3$.

**Corollary 3.** The 3D reconstruction of the centers of multiple spheres using a single image of the spheres from a calibrated camera is possible up to a similarity ambiguity (arbitrary scale) if and only if the real-world ratio between the radii of the spheres is known.

**Proof.** The proof follows from eq. 8 and eq. 9, where the center and radius are correlated; hence, in the case of $n \geq 2$ spheres, the center of the spheres can be determined with consistent scale (i.e., similarity reconstruction) if and only if $n - 1$ unique real-world ratios between the spheres' radii are known a-priori.

Corollary 1 and Corollary 2 show that real-world information from either the radius or center of a sphere is required to recover the sphere's equation in the camera's coordinate frame using a single calibrated image. When only one sphere is present in the image's field of view, Proposition 1 suggests that the equation of one sphere is recoverable up to an arbitrary scale. For multiple spheres captured from a single image, however, Corollary 3 shows that the similarity reconstruction is only possible if the relative radii between the spheres in real-world are known a-priori. Ideally, automated digital documentation of spherical objects warrants a process that requires minimal a-priori knowledge from the real-world. In the following, the recovery of the sphere's equation using multiple image views without a-priori information is discussed.

### 2.1.3 Recovery of sphere equation in multi-view imagery

Given $n \geq 2$ images from a sphere, this sub-section provides a process to calculate the original sphere's center and radius up to a scale ambiguity. This is accomplished by, first, determining the projected sphere's center, $C_C$, in each image independently, and second, triangulating [9] the ellipse centers to determine the 3D center of the sphere, $C$. The calculated center of the sphere is then subjected to rigid body transformation to find the center in each image's coordinate frame. The latter transformed center is used within eq. 10 to calculate the radius of the sphere. In the presence of random measurement errors, the final radius -using least squares minimization- turns out to be the average of the radii obtained from each image.

The projected sphere's center in each image, $C_C$, is calculated by substituting eq. 8 and eq. 9 into eq. 2, where $C_C$ is derived as the function of the projected ellipse parameters and the camera's IOPs as follows:

$$C_C = \frac{1}{f^2+b_e^2}\begin{bmatrix} f^2.x_{Ce} + b_e^2.p_x \\ f^2.y_{Ce} + b_e^2.p_y \end{bmatrix} \tag{11}$$

Using at least two images of the sphere with known projective matrices -see eq. 1- the center of the sphere $C_W$ in world coordinates is determined up to a similarity ambiguity using triangulation [9]. Given that the equations provided in the previous sub-section were derived in the camera coordinate frame, the following rigid body transformation is applied to $C_W$ to derive the sphere's center in each image's coordinate frame:

$$\begin{cases} P_{J_i} = K[Rot_i|t_i] \\ C_i = Rot_i C_W + t_i \end{cases} \tag{12}$$

where $P_{J_i}$, $Rot_i$, and $t_i$, are the projective matrix, rotation, and translation for image $i$ ($i \in \{1 \dots n\}$), respectively, and $C_i = (X_{Ci}, Y_{Ci}, Z_{Ci})$ is the sphere's center in the coordinate frame of image $i$. The radius in each image frame $i$ is then estimated using eq. 10:

$$R_i = \frac{Z_{Ci}.b_{ei}}{\sqrt{b_{ei}^2+f^2}} \tag{13}$$

where $b_{ei}$ is the semi-minor length of the ellipse corresponding to the sphere in image frame $i$, and $R_i$ is the radius of the sphere using the projected ellipse in image frame $i$. An over determined system of equations exists when more than one image frame is considered ($n \geq 2$). In the absence of measurement errors, the radius of the sphere, $R_W$, can be calculated deterministically using only one of the image frames via eq. 13. In the presence of random measurement errors, however, the optimal solution for the overdetermined system can be estimated using least squares adjustment, which becomes the average of the radii obtained from eq. 13:

$$LS(R) = \sum_{i=1}^{n}\left(R - \frac{Z_{Ci}.b_{ei}}{\sqrt{b_{ei}^2+f^2}}\right)^2 \rightarrow \arg\min_R LS(R) : \frac{\partial LS(R)}{\partial R} = 0 \rightarrow \quad R_W = \frac{1}{n}\sum_{i=1}^{n}\frac{Z_{Ci}.b_{ei}}{\sqrt{b_{ei}^2+f^2}} = \frac{1}{n}\sum_{i=1}^{n} R_i \tag{14}$$

where $LS(R)$ is the least squares function, and $\frac{\partial LS(R)}{\partial R}$ represents the partial derivative of $LS(R)$ with respect to variable $R$. In case each observation is assigned a different weight (e.g., weighted based on measurement precision of the estimated image EOPs), the solution to eq. 14 becomes the weighted average of the radii estimated from each image. The process is summarized in Algorithm 1.

---

ALGORITHM 1: Equation of Sphere from Multi-view Imagery

---

**Inputs:** Cameras' projective matrices $P_{J_i}$ ($i \in 1 \dots n$), best fit geometric parameters of the ellipse corresponding to the same sphere in each image, $(x_e, y_e, a_e, b_e, \theta_e)_i$.
**Output:** Parameters of the parent sphere: i.e., center, $C_W$, and radius, $R_W$.
for each image in inputs, do
    calculate the projected center of the sphere in each image, $C_{C_i}$, using eq. 11
end
triangulate the centers, $C_{C_i}$, using the camera's projective matrices to determine the center of the sphere, $C_W$
for each image in inputs, do
    calculate $C_i = (X_{Ci}, Y_{Ci}, Z_{Ci})$ by transforming $C_W$ to the camera's coordinate frame using eq. 12
    calculate the radius of the parent sphere for image $i$, $R_i$, by substituting $Z_{Ci}$ into eq. 13.
end
calculate the average of $R_i$ (eq. 14) to determine the radius of the parent sphere, $R_W$

---

### 2.1.4 Metric scale definition using one sphere with known radius

Using Algorithm 1, the spheres' parameters (center and radius) are calculated up to a scale ambiguity. In case of multiple spheres, the scale remains consistent between different spheres, unlike the case with a single image (see Corollary 3). Hence, if the real-world radius of only one of the spheres in the scene is known a-priori, the equations of all spheres can be determined in real-world scale. This provides the opportunity to add as few as one spherical object (e.g., target) with a known radius into the scene to determine the equation of all spheres in real scale. In other words, if the real-world radius of the spherical object with 3D reconstructed radius of $R_W$ (from Algorithm 1) is $R_R$, the metric scale factor becomes $s_R = \frac{R_R}{R_W}$. In case the ground truth radii of multiple spheres $(n \geq 2)$ are known, the scale in the over-determined system of equations -in terms of least-squares minimization- can be calculated using the following equation, adopted from [10,24]:

$$s_R = \left( \frac{\sum_{i=1}^{n} R_{R_i}^{2}}{\sum_{i=1}^{n} R_{W_i}^{2}} \right)^{\frac{1}{2}} \quad (15)$$

where $R_{R_i}$ and $R_{W_i}$ represent the real-world and estimated radius (Algorithm 1) for sphere $i$. The metric scale, $s_R$, can now be applied to both the centers and radii of the remaining spheres in the scene to determine the sphere's equations in real-world scale.

Algorithm 1 requires the knowledge of: (i) the cameras' projective matrices (i.e., the IOPs and EOPs); (ii) the ellipses corresponding to spheres in each image (spherical ellipses); and (iii) the matching of ellipses corresponding to the same sphere between images. In the following sections, the strategies to automatically derive each of the required inputs are discussed and treated.

## 2.2 Automatic as-built models of spherical structures

Given a set of $n \geq 2$ images acquired from a spherical structure, the proposed solution for automatic generation of as-built digital documentation using images consists of the following steps (Figure 4):

1. Find the EOPs and IOPs of the images (Figure 4a).
2. Select the best pair amongst all images.
3. Detect the spherical ellipses in the image pair (Figure 4b).
4. Match the corresponding ellipses (Figure 4c).
5. For each matched ellipse, perform Algorithm 1 to estimate the center and radius of the parent sphere (Figure 4d).

Here, the proposed solution is designed to only use two out of $n$ images. It is also possible to use all $n$ images to reconstruct the sphere; however, this requires ellipse detection and matching between all images, which may become time consuming for larger number of images (as will be discussed in Section 4.1). Each step is explained in more detail in the following.

### 2.2.1 Deriving Camera EOPs and IOPs

Given a set of overlapping images subjected to rigid body motion (rotation and translation), the sequential SfM process is commonly adopted to automatically estimate both the EOPs and IOPs of the images (Figure 2a and Figure 4a) [2,24]. In this study, COLMAP, a reliable and open-source sequential SfM software was used to perform the coarse reconstruction [35].

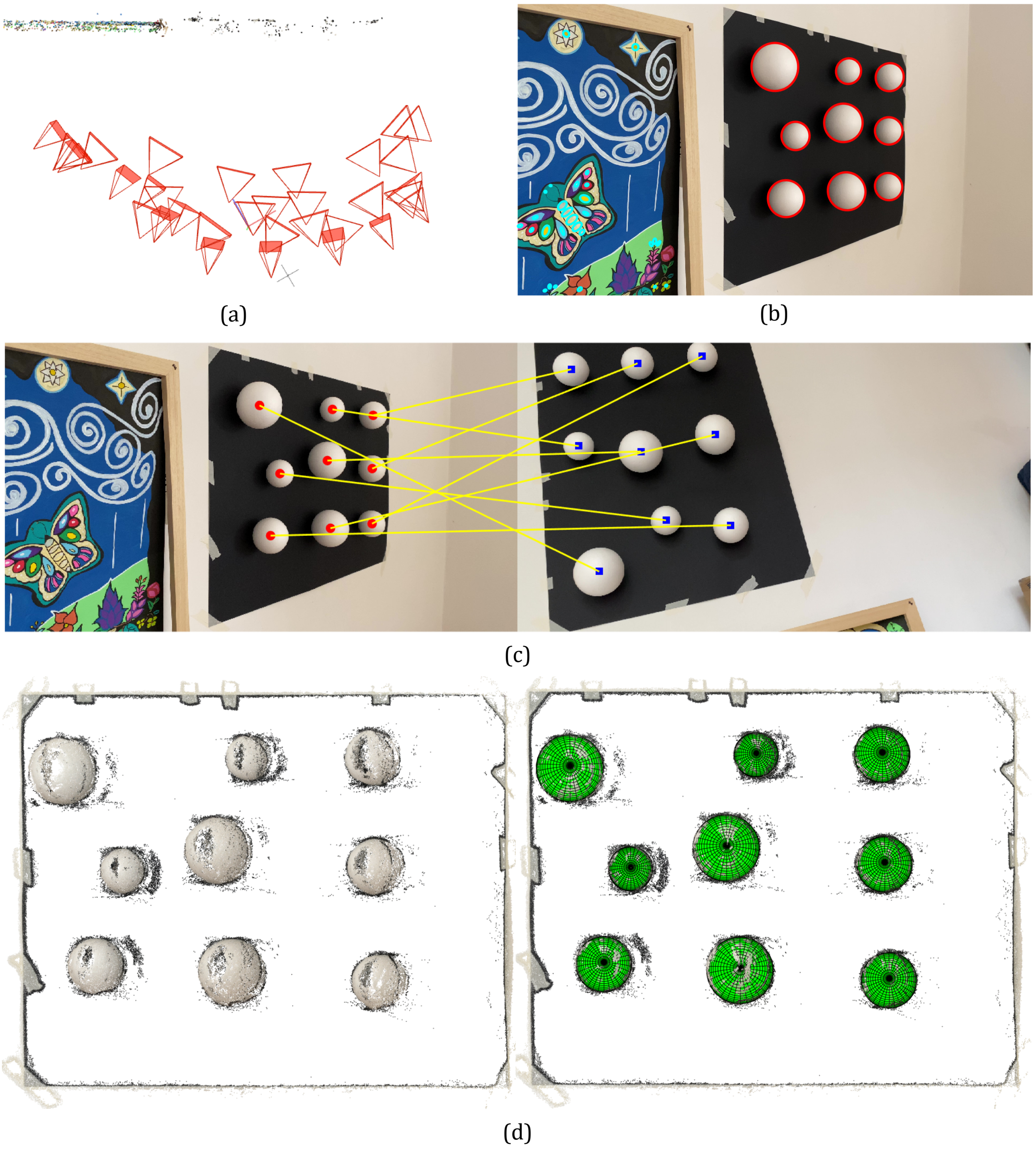

**Figure 4: Visual representation of the methodology: a) coarse 3D reconstruction; b) detected spherical ellipses (red) and eliminated remaining detected ellipses (cyan); c) matched spherical ellipses between best pair of images; d) as-built model of the spheres superimposed onto the dense 3D point cloud.**

### 2.2.2 Selecting the best pair of images in the network

Two established metrics were utilized to select the best pair of images. The first is the average convergence angle between two images, $\alpha_{ij}$, introduced in [35]. The second metric is the network overlap of an image, $Ov_i$, introduced in [24]. Best practices suggest that both metrics must be as large as possible to provide accurate 3D reconstruction. To account for this, the objective function, $\vartheta_{ij}$, between images $i$ and $j$ is defined:

$$\vartheta_{ij} = \frac{\alpha_{ij}}{\alpha_{max}} + \frac{Ov_i + Ov_j}{2.Ov_{max}} \tag{16}$$

where $\alpha_{ij}$ is the average of the convergence angle between two images $i$ and $j$, $\alpha_{max}$ is the maximum of the convergence angle between all pair of images, $Ov_i$ is the network overlap of image $i$ and $Ov_{max}$ is the maximum network overlap in all images. The pair of images achieving the maximum of $\vartheta_{ij}$ are chosen as the best pair. Since the best practices also suggest that the average convergence angle between images must be greater than 20° [35], only pair of images with $\alpha_{ij}$ >20° will be considered.

**2.2.3 Detecting spherical ellipses**

Detecting spherical ellipses requires: (i) an efficient method for ellipse detection from images; and (ii) a reliable procedure for elimination of ellipses that do not represent projections of spherical objects. To detect ellipses from images, the robust ellipse detection method of [22] is utilized. To eliminate the unwanted ellipses (cyan coloured ellipses in Figure 4b), the relationship between the major and minor axis of the spherical ellipses, derived in Section 3.1.2, are exploited. This relationship can be derived in closed form by substituting eq. 8 and eq. 9 into eq. 6:

$$a_e = b_e \sqrt{\frac{(x_{Ce} - p_x)^2 + (y_{Ce} - p_y)^2}{f^2 + b_e^2} + 1} \tag{17}$$

Therefore, any ellipse which represents a projection of spheres onto an image must follow the relationship derived in eq. 17. In the absence of random measurement errors, the equality is exact. In practical settings with random measurement errors, however, a margin of error exists. This margin can, however, be analytically estimated to a first order error term using the law of variance propagation [13]. As such, for each detected ellipse the following equations are calculated:

$$\begin{cases} \tau = 1 - \frac{b_e}{a_e}\sqrt{\frac{(x_{Ce} - p_x)^2 + (y_{Ce} - p_y)^2}{f^2 + b_e^2} + 1} \\ \sigma_\tau^2 = J_\tau\, \Sigma^{var}\, J_\tau^T, \quad J_\tau = \begin{bmatrix} \frac{\partial\tau}{\partial a_e} & \frac{\partial\tau}{\partial b_e} & \frac{\partial\tau}{\partial x_{Ce}} & \frac{\partial\tau}{\partial y_{Ce}} & \frac{\partial\tau}{\partial p_x} & \frac{\partial\tau}{\partial p_y} & \frac{\partial\tau}{\partial f} \end{bmatrix}, \quad {\Sigma^{var}}_{7\times7} = \begin{bmatrix} {\Sigma^{Ellipse}}_{4\times4} & 0_{4\times3} \\ 0_{3\times4} & {\Sigma^{IOP}}_{3\times3} \end{bmatrix} \end{cases} \tag{18}$$

where $J_\tau$ is the Jacobian matrix of $\tau$ with respect to the relational variables, $\frac{\partial\tau}{\partial A}$ is the partial derivative of function $\tau$ with respect to variable $A$, $\Sigma^{var}$ is the covariance matrix of the variables, $\Sigma^{Ellipse}$ and $\Sigma^{IOP}$ are the covariance matrices of the best fit ellipse parameters $(a_e, b_e, x_{Ce}, y_{Ce})$, and IOPs $(p_x, p_y, f)$, respectively, and $\sigma_\tau$ is the standard deviation of the defined function $\tau$. For spherical ellipses, according to eq. 17, $\tau = 0$. To accommodate for the random measurement errors, the following condition is used:

$$\begin{cases} |\tau| \le k.\sigma_\tau, & \text{Spherical ellipse} \\ |\tau| > k.\sigma_\tau, & \text{Undesirable ellipse} \end{cases} \tag{19}$$

In this study, $k = 2$ is used to accommodate for 95% confidence under the normality assumption. To estimate $\sigma_\tau$ in eq. 18, the covariance of the geometric parameters of the ellipse, $\Sigma^{Ellipse}$, are estimated using Algorithm 4 of [37], and the covariance of the camera's IOPs, ${\Sigma^{IOP}}_{3\times3}$, are derived from the bundle adjustment during the SfM process [20]. Here, since the IOPs are estimated within the SfM process (see Section 2.2.1) and the ellipse geometric parameters are estimated in each image independent from the IOPs, their variances are considered uncorrelated. The latter is represented by the zero matrices within $\Sigma^{var}$, eq. 18. The closed formulation for $J_\tau$ is provided in Appendix A.

### 2.2.4 Matching spherical ellipses

Once the spherical ellipses of the selected pair of images are automatically detected, the ellipses corresponding to the same sphere must be matched. Therefore, a two-step process is devised. For each ellipse in the first image, first, the ellipses in the second image, whose corrected center satisfy the epipolar [9] constraint are recognized. Here, the parameter-less epipolar constraint presented in Algorithm 2 (step 4) of [20] is adopted. The ellipse in the first image is then used together with each of the possible matching candidates from the second image to estimate the center and radius using Algorithm 1. The estimated sphere is then used within eq. 6 and eq. 7 to estimate the reprojected geometric parameters of the ellipses. The ellipse whose distance between the original ellipses and the corresponding reprojected geometric parameters are the smallest is considered as the matching spherical ellipse. The process is summarized in Algorithm 2.

ALGORITHM 2: Information Modeling of Spheres using Images

**Inputs:** Set of $n$ overlaping images from spherical objects.
**Output:** Radius and center of the parent spheres.

```
perform structure-from-motion and calculations
    for each image i = 1:n in inputs, do
        determine scale invariant features using SIFT
        calculate the network overlap, Ov_i
    end
    for images i = 1:n − 1 in inputs, do
        for images j = 2:n in inputs, do
            match determined features between images i and j
            perform sequential reconstruction and bundle adjustment and estimate the EOPs and IOPs
            calculate average convergence angle between the two images, α_ij
            calculate the objective function ϑ_ij (eq. 16)
        end
    end
    find the two images, l and k which maximize ϑ_ij in eq. 16
for each of the selected images l and k, do
    detect spherical ellipses using the robust ellipse detection and fitting method of [21,22]
    retain only ellipses satisfying the sphere condition of eq. 19
end
for each ellipse in image l, do
    for each ellipse in image k, do
        estimate the radius and center of the hypothesized parent sphere using Algorithm 1
        rotate the center of the estimated sphere into each images' coordinate frame using eq. 12
        estimate the parameters of the reprojected ellipses from the sphere parameters using eq. 6 and eq. 7
        calculate the distance between the parameters of the original ellipses from those of the reprojection
        select the matching ellipse as that obtaining the smallest ellipse parameter distance
    end
end
```

# 3 Experimental design and calculation

Two sets of experiments under both laboratory and real-world conditions were designed. The first was the laboratory experiment, which investigated the effectiveness of the proposed method in detection and 3D modeling of spherical black and white targets. The second consisted of two sets of the real-world experiments, which verified the effectiveness of (i) the method for metric scale definition of photogrammetric point clouds of a heritage column; and (ii) as-built modeling of two separate heritage hemispherical domes. The summary of the designed experiments, along with the type of data and the experiment's purpose are given in Table 1. The data related to the experiments are also demonstrated in Figure 5 and will be explained in more detail in the following.

**Table 1:** **Summary of the designed experiments**

| Experiment Description | Type of data | Instrument | Purpose |
|---|---|---|---|
| Laboratory experiment | Thirty images from nine spherical Styrofoam ball in the laboratory | iPhone 11 | Evaluating the effectiveness of Algorithm 2 as the number of image views increase under laboratory conditions |
| Real-world experiment: scale definition using spherical targets | Thirty images from two spherical Styrofoam ball and one heritage column at the Großherzogliche Grabkapelle cathedral in Karlsruhe, Germany | iPhone 11 | Assessing the feasibility and accuracy of utilizing the proposed method of Algorithm 2 and eq. 15 for metric scale definition of photogrammetric networks |
| Real-world experiment: as-built modelling of hemispherical domes | 100 images from the heritage hemispherical dome at the Orangerie in Karlsruhe, Germany | iPhone 11 | Evaluating the accuracy and computation time for the proposed method of Algorithm 2, compared to current best practice using point cloud analysis |
| | 66 images of one heritage hemispherical dome at the Chunakhola Masjid in Bagerhat, Bangladesh | DJI Phantom 4 Pro drone | |

## 3.1 Method of validation of results

In each experiment, the effectiveness of the radius and center estimated using Algorithm 2 were compared to a ground truth. The following generic steps were taken for each experiment to validate the results:

1. Ground truth radius and center estimation:
    1.1. Perform dense 3D reconstruction (Figure 2b) using COLMAP.
    1.2. Detect the spherical objects from the 3D point clouds using the robust method of [23].
2. Calculate the percentage of root mean squared error (P-RMSE) as follows:

$$\text{P-RMSE}(\%) = \frac{RMSE}{R} . 100 \qquad (20)$$

where $RMSE$ is the root mean squared error between the estimated (Algorithm 2) vs. the ground truth of the sphere's parameters (radius and center), and $R$ is the ground truth radius.[1] The P-RMSE is adopted here to normalize the accuracy results for datasets with domes of different radii.

3. Compare the computation time between Algorithm 2 and the ground truth.

Some of the experiments also contained specific steps for the analysis of the results which will be discussed in more detail in the following.

[1] In case the base of the measurement is a distance instead of a radius (such as the case in the scale definition experiment), the ground truth radius is replaced by the ground truth distance.

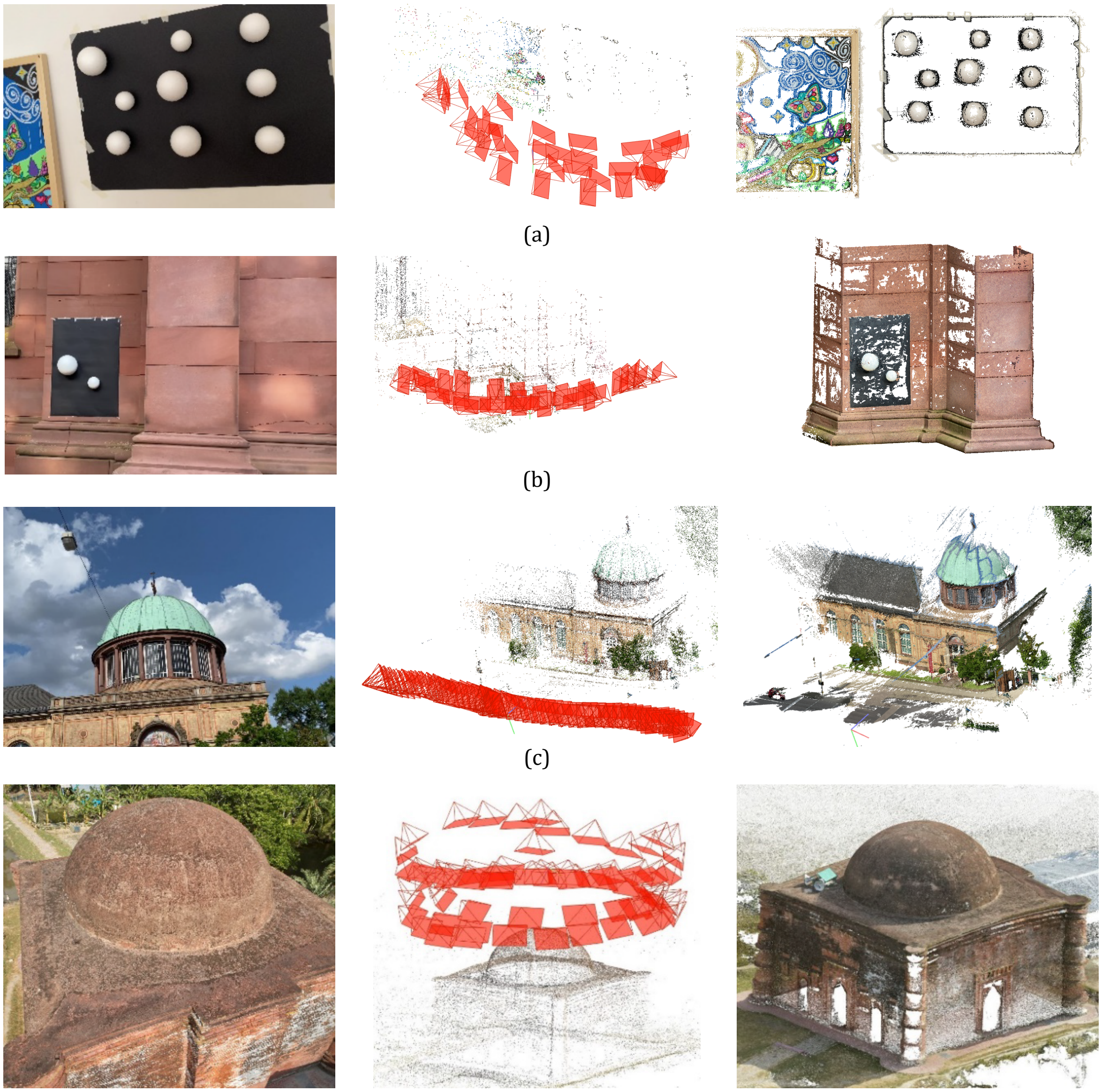

(a)

(b)

(c)

(d)

**Figure 5: Sample image (left), coarse 3D reconstruction (middle), and dense 3D reconstruction (right) for: a) laboratory experiment; b) Orangerie dome (Karlsruhe); and c) Chunakhola Masjid dome (Bagerhat).**

## 3.2 Laboratory experiment

The objective of this experiment was to validate the correctness of the method for sphere detection and modeling, presented in Algorithm 2, as the number of image views increased. For this, nine white spherical Styrofoam balls, attached onto a black cardboard background, were used as the subject of the experiment (Figure 5a-left). Thirty

(30) 4K images using the iPhone 11 were acquired from the targets (Figure 5a-middle). The impact of increasing the number of images on the P-RMSE of the center and radius of the parent spheres was examined. Given that Algorithm 2 only considers the best pair of images, to accommodate for the increasing number of image views, the best pair is replaced by a random set of $k = 2 \ldots 30$ images using the following steps:

1. Calculate the number of simulations, $p = \min\left(50, \binom{30}{k}\right)$.
2. Randomly select $p$ unique sets of $k$ images out of 30 images.
3. For each set of $k$ images, perform Algorithm 2.
4. Calculate the accuracy of the center and radius (P-RMSE; Section 4.1).
5. Record the mean, maximum and minimum of the P-RMSE of all $p$ sets of $k$ images.

## 3.3 Real-world experiments

### 3.3.1 Scale definition using spherical targets

The goal of this experiment was to verify the effectiveness of the proposed method for automatic metric scale definition of photogrammetric networks using spherical targets (Algorithm 2 together with Section 2.1.4). One dataset comprised of thirty (30) 4K iPhone 11 images were taken from one of the external columns of the Großherzogliche Grabkapelle cathedral in Karlsruhe, Germany. Two spherical Styrofoam balls similar to the previous laboratory experiment were used to define metric scale for the network. Using the estimated scale, the RMSE of the dimensions of the column in the photogrammetric point cloud with respect to the ground truth dimensions were reported, and the following steps were performed:

1. Ground truth pre-processing:
    - 1.1. Perform the robust column detection of Maalek [25] on collected laser scanner point clouds, acquired using the Leica BLK360. The robust rectangular column of [25] is comprised of the robust planar classification and segmentation of Maalek [17], followed by a rule-based semantic column detection based on the direction of the majority columns' plane normal vectors.
    - 1.2. Model the rectangular column [25], and record the planar patches for the three observable sides (cyan coloured patches in Figure 6b).
2. Scale definition:
    - 2.1. Perform Algorithm 2 to automatically model and estimate the radius of the spherical targets. These targets are shown with red and green balls in Figure 6b.
    - 2.2. Estimate the metric scale using eq. 15 using the ground truth radii.
    - 2.3. Scale the dense 3D reconstructed photogrammetric point cloud.
3. Perform the point cloud vs. BIM method of [19] on the modeled column and the scaled photogrammetric point cloud and record the RMSE.

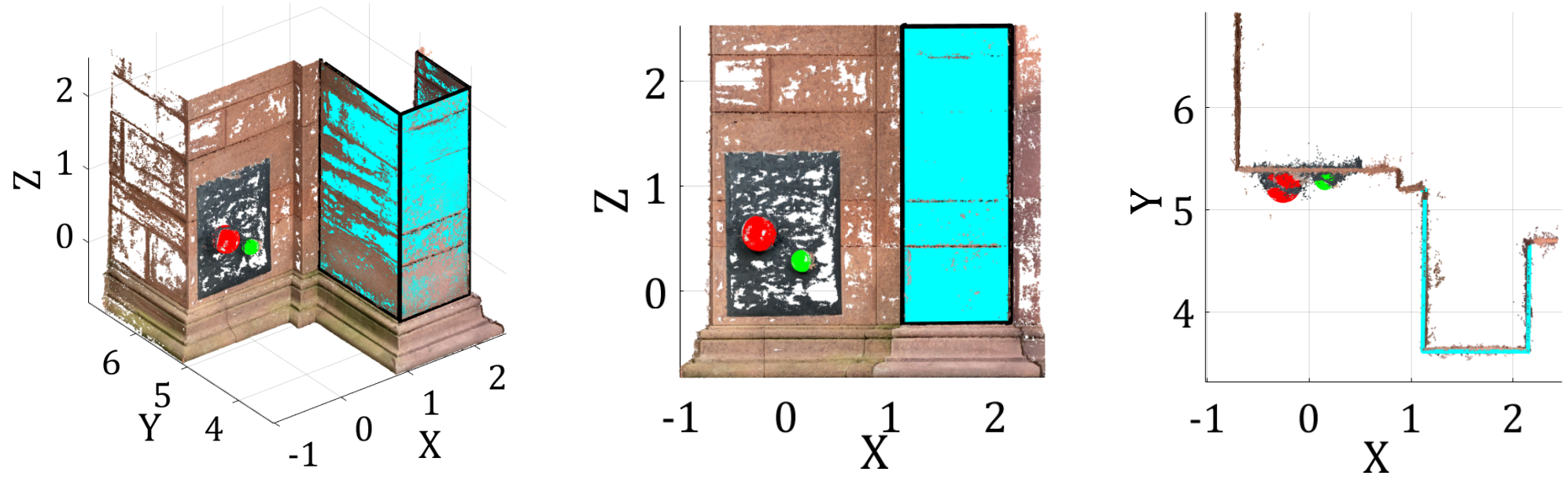

**Figure 6:** As-built model of the column and spherical targets (subjects of the scale definition experiment) superimposed onto the original 3D reconstructed photogrammetric point cloud in 3D orthographic view (left), front view (middle) and top view (right).

### 3.3.2 As-built modelling of hemispherical domes

Two datasets from hemispherical heritage domes were used in this study. The first dataset consisted of 100 4K images of the Orangerie in Karlsruhe, Germany (Figure 5c), acquired using the iPhone 11. The second dataset included 66 images taken from the Chunakhola Masjid in Bagerhat, Bangladesh (Figure 5d) using the DJI Phantom 4 Pro drone. The second dataset was collected by CyArk and is publicly available at the Open Heritage 3D project [6]. For both datasets, the P-RMSE and computation time using the best pair of images were reported.

# 4 Experimental results and discussions

## 4.1 Laboratory experiment

Figures 7a and 7b show the results of the parameter estimation accuracy (P-RMSE) and the average computation times as the number of image views increased from 2 to 30, respectively. To provide a fair representation of the impact of number of images on the computation time, Figure 7b considered only the computation time of Algorithm 2 steps 2-5 and excludes the initial IOP/EOP estimation using SfM. Three main observations were made from the results shown in Figure 7. First, Figure 7a shows that the accuracy of the estimated center and radius improved as the number of images increased. The uncertainty region between the minimum and maximum (confidence region) also reduced as the number of image views increased due to the improved redundancy and network geometry [16,20]. The second observation was that the parameter estimation error using the best pair was consistent with the expected value (average) of the errors obtained using eight image views. This particularly signifies the advantage of adopting the proposed best image pair selection criteria (eq. 16) compared to random sampling. The last observation was related to the average computation times shown in Figure 7b. It can be observed that the computation time increased as the number of image views increased. The best pair using Algorithm 2 achieved parameter estimation accuracy consistent with eight randomly selected image views, whereas the average computation time using eight image views was approximately 50 times that using two images.

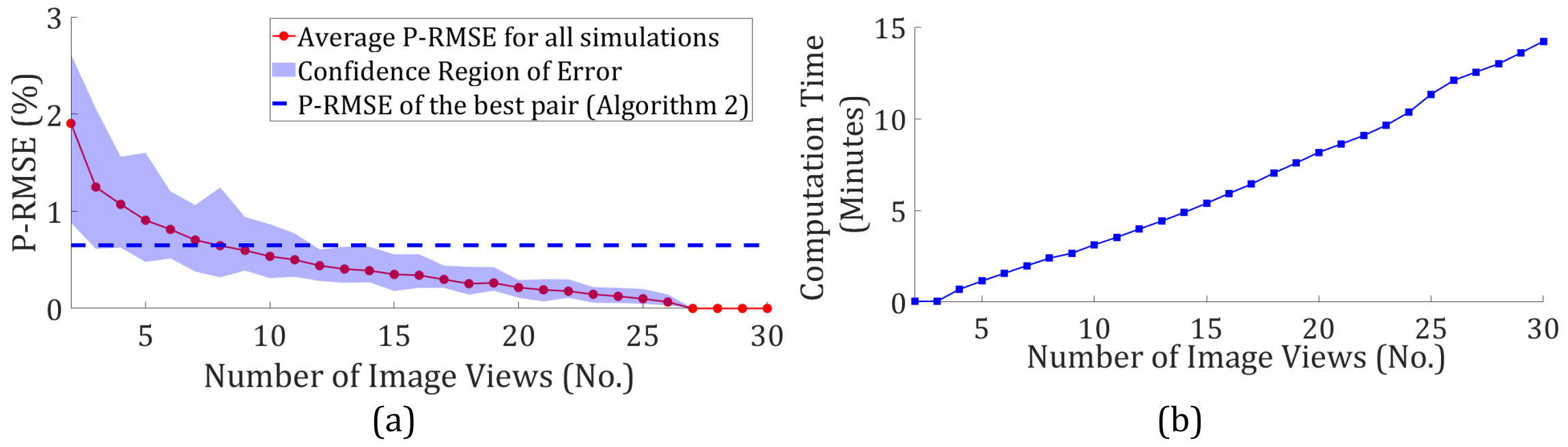

**Figure 7: Results of the laboratory experiment: number of image views vs. a) P-RMSE; and b) computation time.**

## 4.2 Real-world experiments

### 4.2.1 Scale definition using spherical targets

In this experiment, the effectiveness of Algorithm 2 in modeling spherical targets for the purpose of metric scale definition in photogrammetric point clouds was examined. Two spherical black and white targets were used to first define the metric scale automatically. The scale was then used to calculate the RMSE of the heritage column of the Großherzogliche Grabkapelle cathedral in Karlsruhe, Germany. The results of the spherical ellipse detection and final modelling of the spherical targets using Algorithm 2 are shown in Figure 8a and Figure 6, respectively.

The targets are in two different sizes with real-world radii of 10cm and 6cm. Using eq. 15, the real-world radii and the radii obtained using Algorithm 2, the scale of the overdetermined set of radii was determined. The precision of the scale definition (the RMSE of the spheres' radii after applying the scale factor) was 0.03mm.

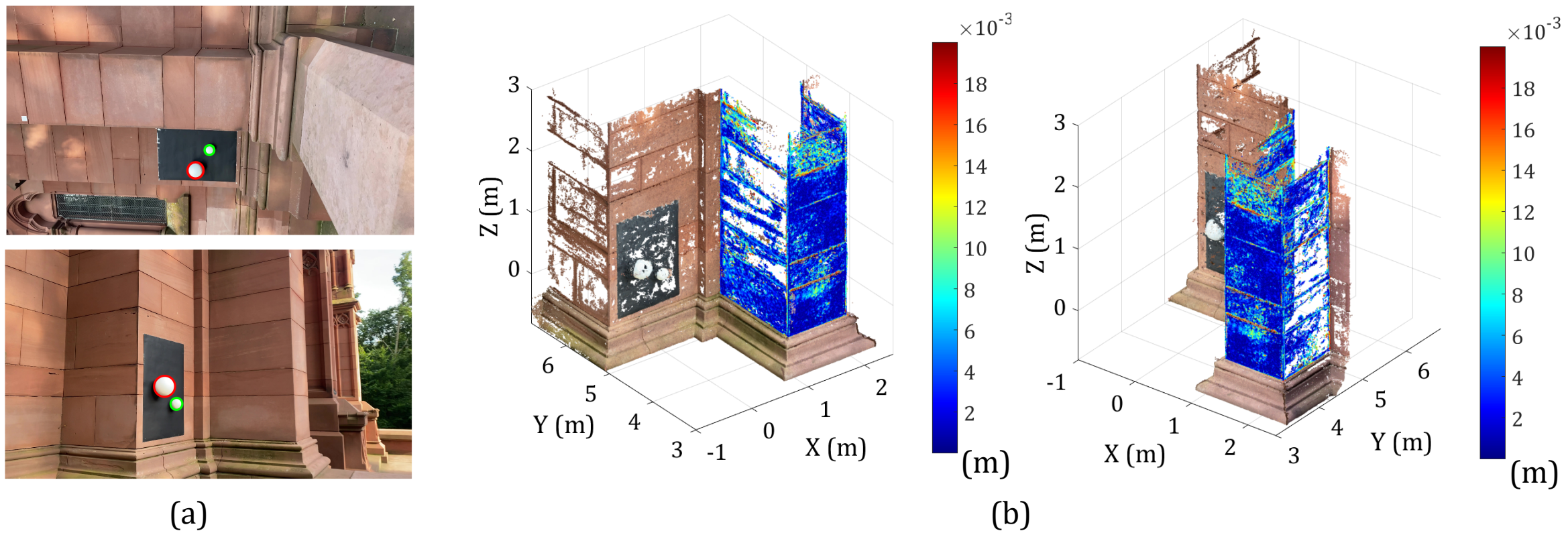

**Figure 8: Real-world scale definition experiment: a) selected best pair of images along with the two detected spherical ellipses; and b) final 3D model of the detected spheres superimposed onto the 3D reconstructed point cloud in orthographic view.**

As described in Section 3.3.1, the scale was applied to the whole point cloud and the RMSE of the point cloud to the automatically generated as-built model was calculated. The accuracy of the automatically generated as-built model (modelled as three planar surfaces as demonstrated in Figure 6) from the laser scanner was around 0.5mm. Figure 8b shows the heatmap of the distance of the scaled inlier photogrammetric point clouds to the as-built model of the heritage column. The RMSE of the photogrammetric point clouds from the model (the average of the distances observed in Figure 8b) was 3.1mm, which is consistent with the precision of photogrammetric measurements [24].

### 4.2.2 As-built modelling of hemispherical domes

Figures 9 and 10 illustrate the results of the best selected pair of images along with the modeled sphere, corresponding to the hemispherical domes at the Orangerie (Karlsruhe), and Chunakhola Masjid (Bagerhat), respectively. The P-RMSE of the estimated spherical dome's parameters (compared to ground truth parameters) were 0.64% and 0.61% for the Orangerie and the Masjid, respectively. Consistent with the laboratory experiments, the results suggest that if the ground truth radius of the domes were 1m, the center and radius of the dome are estimated with 6.4mm and 6.1mm accuracy for the Orangerie and the Masjid, respectively.

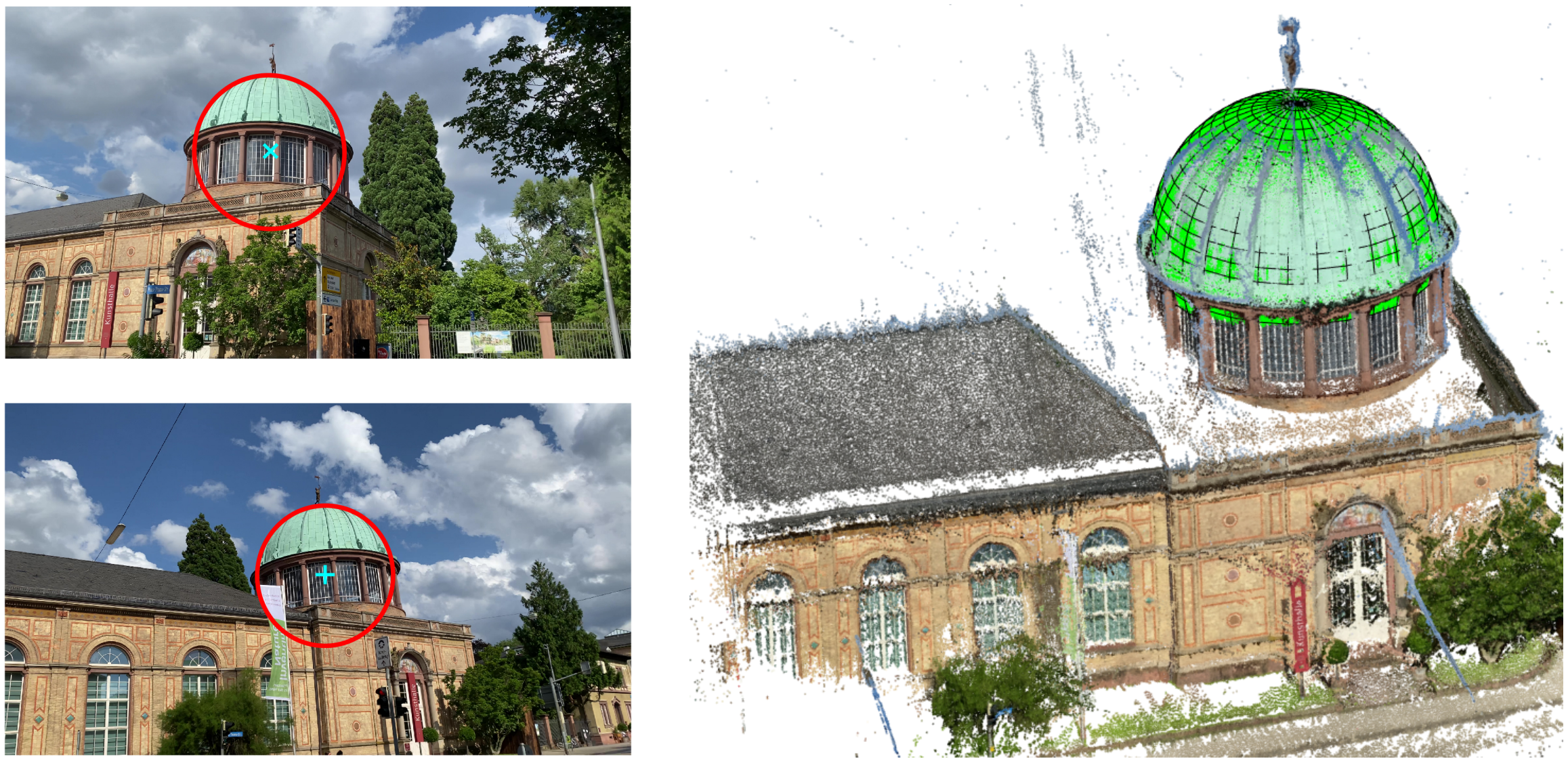

(a) (b)

**Figure 9: Results of as-built modeling for the Orangerie in Karlsruhe using Algorithm 2: a) detected and matched best pair of images; and b) generated sphere superimposed onto the dense 3D reconstruction point cloud.**

Table 2 shows the results of the computation time using Algorithm 2 and the ground truth method (step 1 of Section 4.1). It can be observed that for both real-world hemispherical domes, Algorithm 2 was approximately 7 times faster. To provide some perspective, around 9 domes of the Orangerie in Karlsruhe can be automatically modelled in one day using Algorithm 2, whereas the same 9 domes will take approximately a week using the current best practices. We conclude that the proposed method is promising particularly for the large-scale monitoring and modeling of hemispherical domes.

**Table 2: Summary of the computation times using Algorithm 2 vs. the ground truth**

| **Method** | **Orangerie in Karlsruhe** | **Chunakhola Majid in Bagerhat Purpose** |
|---|---|---|
| Ground truth: Dense 3D reconstruction + Sphere detection [35] | 191.017 | 134.174 |
| Proposed: Coarse 3D reconstruction + Algorithm 2 | 26.378 | 17.837 |

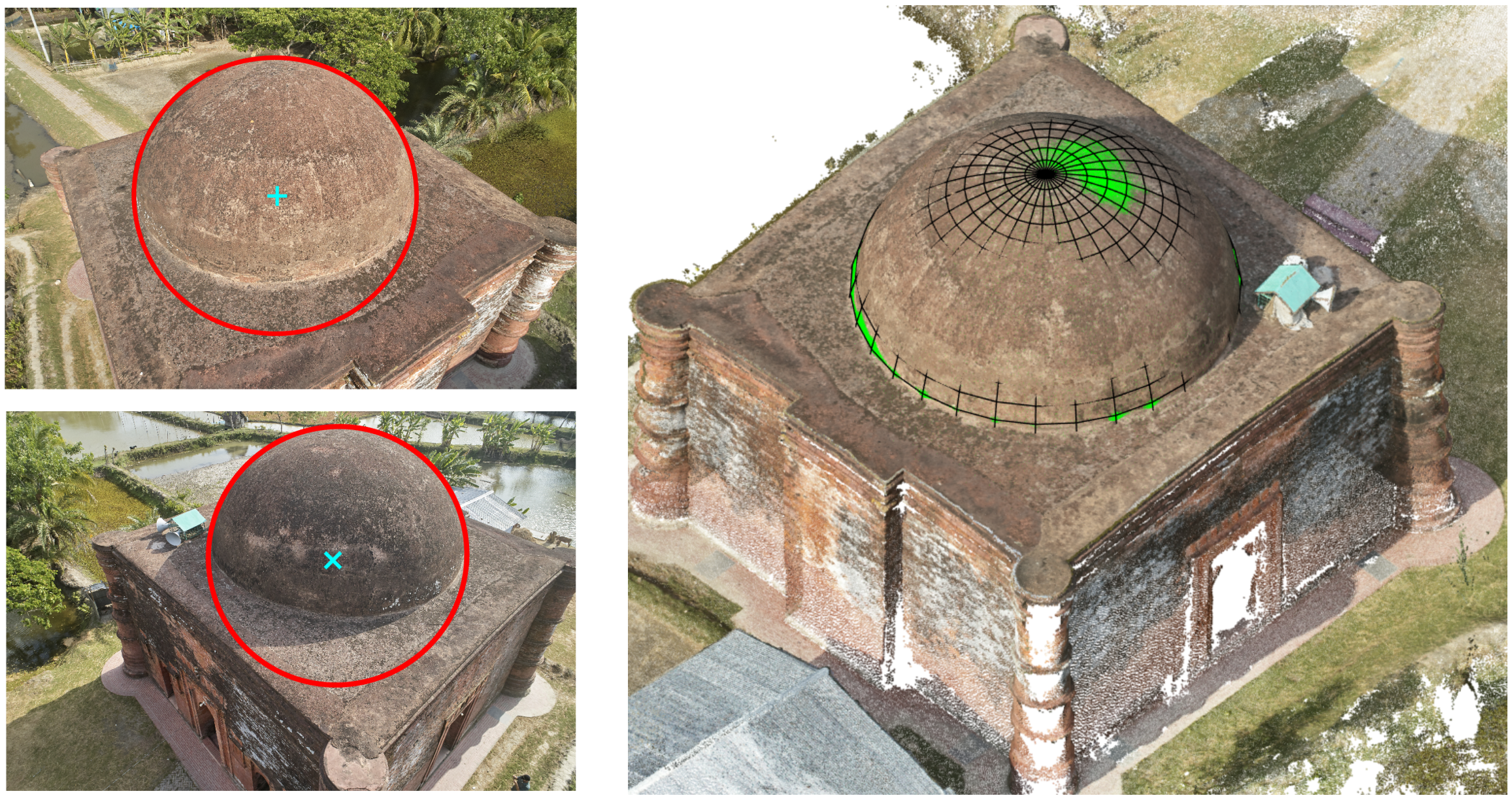

(a) (b)

**Figure 10: Results of as-built modeling for the Chunakhola in Bagerhat using Algorithm 2: a) detected and matched best pair of images; and b) generated sphere superimposed onto the dense 3D reconstruction point cloud.**

# 5 Conclusions

A new method for detecting and automatic modeling of spherical objects, applicable to as-built model generation of hemispherical domes, was presented. To this end, the closed-form solutions to the projection of spherical objects onto perspective cameras were derived. Using these original formulations, the proposed method only required two images from a network of images to provide accurate parameter estimation of the parent sphere, while maintaining computational efficiency. The effectiveness of the proposed method was investigated under both laboratory and real-world conditions. It was observed that the proposed method, which uses a selection criterion to choose the best pair of images, performed as good as the case when eight random images were selected, while achieving around 50 times faster computation time.

In the real-world experiments, the proposed method for scale definition using spherical targets achieved an accuracy of 3.1mm (within the expected measurement uncertainty of the photogrammetric point clouds using smartphones), when compared to modeled heritage columns using laser scanner point clouds. In the digital modeling of spherical domes experiment, it was observed that the proposed method achieved consistent P-RMSE between both real-world and laboratory (0.62% on average). Furthermore, it was revealed that the real-world domes were processed approximately 7 times faster than the current best practice, which utilizes point cloud processing from dense 3D reconstructed point clouds. This latter is a difference between modeling several domes in one day vs. one week. This reduction of time is not only convenient for the timely accessibility of data for the project's stakeholders (e.g., the public, decision makers, ministries, and etc.), but it will also reduce costs and resources related to computational infrastructures (e.g., cloud computing services, servers, and etc.). We conclude that the proposed method provides a practical solution for the large-scale monitoring and digital documentation of spherical objects, such as hemispherical domes. The authors hope that the methodology and results presented in this study will encourage other researchers in the broader domain of construction informatics to search for alternative and simpler approaches to object detection and digital twin generation problems, before resorting to more advanced and computationally expensive point cloud processing methods.

## ACKNOWLEDGMENTS

No funding was received for this project.

# 6 HISTORY DATES

# APPENDIX A: CLOSED-FORM JACOBIAN MATRIX

This section presents the closed formulation for the Jacobian matrix, $J_\tau$, of the function τ, defined in eq. 17:

$$\tau = 1 - \frac{b_e}{a_e}\sqrt{\frac{(x_{Ce}-p_x)^2+\left(y_{Ce}-p_y\right)^2}{f^2+b_e^2}+1} \quad (A1)$$

$$J_\tau = \left[\frac{\partial\tau}{\partial a_e} \quad \frac{\partial\tau}{\partial b_e} \quad \frac{\partial\tau}{\partial x_{Ce}} \quad \frac{\partial\tau}{\partial y_{Ce}} \quad \frac{\partial\tau}{\partial p_x} \quad \frac{\partial\tau}{\partial p_y} \quad \frac{\partial\tau}{\partial f}\right] \quad (A2)$$

$$J_\tau : \begin{cases} \frac{\partial\tau}{\partial a_e} = \frac{1-\tau}{a_e} \\ \frac{\partial\tau}{\partial b_e} = -\frac{f^2(1-\tau)}{b_e(f^2+b_e^2)} - \frac{b_e^3}{a_e^2(1-\tau)(f^2+b_e^2)} \\ \frac{\partial\tau}{\partial x_{Ce}} = -\frac{b_e^2(x_{Ce}-p_x)}{a_e^2(1-\tau)(f^2+b_e^2)} \\ \frac{\partial\tau}{\partial y_{Ce}} = -\frac{b_e^2(y_{Ce}-p_y)}{a_e^2(1-\tau)(f^2+b_e^2)} \\ \frac{\partial\tau}{\partial p_x} = \frac{b_e^2(x_{Ce}-p_x)}{a_e^2(1-\tau)(f^2+b_e^2)} \\ \frac{\partial\tau}{\partial p_y} = \frac{b_e^2(y_{Ce}-p_y)}{a_e^2(1-\tau)(f^2+b_e^2)} \\ \frac{\partial\tau}{\partial f} = \frac{f}{a_e^2(1-\tau)(f^2+b_e^2)}\left(a_e^2(1-\tau)^2 - b_e^2\right) \end{cases} \quad (A3)$$